\documentclass[sn-apa]{sn-jnl}

\usepackage{graphicx}%
\usepackage{multirow}%
\usepackage{amsmath,amssymb,amsfonts}%
\usepackage{amsthm}%
\usepackage{mathrsfs}%
\usepackage{xcolor}%
\usepackage{textcomp}%
\usepackage{booktabs}%
\usepackage{listings}%

\usepackage{hyperref}
\usepackage{array,multirow}
\usepackage{tikz}

\usepackage{pgfplots}
\usetikzlibrary {arrows.meta}

\usepackage{tabularx}
\usepackage{tabulary}

\usepackage{arabtex}
\usepackage{utf8}
\setcode{utf8}

\begin{document}

\title[Article Title]{Dhati+: Fine-tuned Large Language Models for Arabic Subjectivity Evaluation}


\author*[1,2]{\fnm{Slimane} \sur{Bellaouar}}\email{bellaouar.slimane@univ-ghardaia.dz}

\author*[3,4]{\fnm{Attia} \sur{Nehar}}\email{neharattia@univ-djelfa.dz}
\equalcont{These authors contributed equally to this work.}

\author[1]{\fnm{Soumia} \sur{Souffi}}\email{soumia.souffi@univ-ghardaia.dz}
\equalcont{These authors contributed equally to this work.}

\author[1]{\fnm{Mounia} \sur{Bouameur}}\email{bouameur.mounia@univ-ghardaia.dz}

\equalcont{These authors contributed equally to this work.}

\affil*[1]{\orgdiv{Dept. of Mathematics and Computer Science}, \orgname{Université de Ghardaia, Algeria}, 
}

\affil[2]{\orgdiv{Lab. des Mathématiques et Sciences Appliquées (LMSA)}, \orgname{Université de Ghardaia, Algeria}
}

\affil*[3]{\orgdiv{Exact Sciences and Computer Science Faculty}, \orgname{Ziane Achour University, Algeria}
}

\affil[4]{\orgdiv{Lab. d'Informatique et Mathématiques (LIM)}, \orgname{Université Amar Telidji, Algeria}
}

\abstract{Despite its significance, Arabic, a linguistically rich and morphologically complex language, faces the challenge of being under-resourced. The scarcity of large annotated datasets hampers the development of accurate tools for subjectivity analysis in Arabic. 
Recent advances in deep learning and Transformers have proven highly effective for text classification in English and French. This paper proposes a new approach for subjectivity assessment in Arabic textual data. To address the dearth of specialized annotated datasets, we developed a comprehensive dataset, AraDhati+, by leveraging existing Arabic datasets and collections (ASTD, LABR, HARD, and SANAD). Subsequently, we fine-tuned state-of-the-art Arabic language models (XLM-RoBERTa, AraBERT, and ArabianGPT) on AraDhati+ for effective subjectivity classification. 
Furthermore, we experimented with an ensemble decision approach to harness the strengths of individual models. 
Our approach achieves a remarkable accuracy of 97.79\,\% for Arabic subjectivity classification. 
Results demonstrate the effectiveness of the proposed 
approach in addressing the challenges posed by limited resources in Arabic language processing.}


\maketitle

\section{Introduction}
\label{sec:intro}


The proliferation of digital technology has resulted in a surge of information and a rise in user-generated content, encompassing various opinions and thoughts. This pattern is anticipated to persist as social media and other online platforms gain wider prevalence. These opinions and thoughts offer valuable perspectives into consumer behavior and public sentiment, thereby rendering sentiment analysis an increasingly vital instrument for businesses, governments, and researchers. In this context, subjectivity classification serves to ascertain whether the content of a text is subjective or objective, constituting a pivotal stage in sentiment analysis.

Arabic is a language of great complexity and richness. It is spoken by over 420 million people globally, securing its position as the fifth most spoken language worldwide. The extensive usage of this language is a testament to its significance as a fundamental aspect of cultural identity in numerous countries and its status as an official language in various governments and international organizations.
Arabic holds a significant position in the digital world, being among the top 10 most used languages on the internet. More than 280 million individuals actively participate in online content in Arabic. Despite the challenging characteristics of the Arabic language, the expansion of the Arabic-speaking internet community has stimulated an increase in the need for sophisticated natural language processing (NLP) technologies specifically designed for Arabic. In particular, there is an increasing interest in creating strong sentiment analysis algorithms that can accurately understand the subtle aspects of sentiment and emotion expressed in Arabic.

Prior research on subjectivity classification in Arabic has predominantly utilized traditional, machine learning, and deep learning approaches. Traditional methods utilize manually created features such as lexicon-based and morphological-based features \citep{Awwad2016PerformanceCO,abdul-mageed-etal-2011-subjectivity,abdul2014samar}. However, in the realm of machine learning methods, only three classifiers have consistently demonstrated extraordinary outcomes: The paper by \cite{OUESLATI2020408review} reviews the Support Vector Machine (SVM), k-Nearest Neighbor (KNN), and Naive Bayes (NB) algorithms.

 
Recent progress in deep learning has led to a significant adoption of  Recurrent Neural Networks (RNNs) \citep{Alhumoud2022}, ensemble methods \citep{Alharbi2021,Mohamed2022}, and fine-tuning pre-trained language models \citep{Alduailej2022}. These methodologies have gained popularity because of their capacity to catch intricate patterns in data, especially in jobs related to natural language processing and text classification.


Building upon our previous work on the Dhati tool for Arabic subjectivity analysis \citep{nehar2023dhati}, this study aims to enhance its capabilities. To achieve this, we propose a novel approach involving the fine-tuning of three Large Language Models (LLMs) on our expanded dataset. This augmented dataset, henceforth referred to as AraDhati+, incorporates the Arabic Sentiment Tweets Dataset (ASTD) \citep{nabil-etal-2015-astd}, the Large Arabic Books Reviews (LABR) \citep{aly-atiya-2013-labr}, the Hotel Arabic-Review Dataset (HARD) \citep{ELNAGAR2018182AnAnnotated}, and the Single-labeled Arabic News Articles Dataset (SANAD) \citep{einea2019sanad}.

The subsequent sections of this paper are organized in the following manner: Section~\ref{sec:prelim} offers the essential background and fundamental knowledge required to understand the following parts. Section~\ref{sec:related} examines the studies that have been conducted on sentiment and subjectivity assessment. We provide a detailed explanation of our proposed approach for evaluating Arabic subjectivity in Section \ref{sec:method}. Section \ref{sec:experiments} presents the findings of the experiments and includes a discussion of these results. Ultimately, we derive conclusions and outline future endeavors in section \ref{sec:conc}.

\section{Opinion, Sentiment, and Subjectivity}
\label{sec:prelim}


In the field of textual information analysis, the concepts of opinion, sentiment, and subjectivity are essential to understanding the nuances of human communication. The purpose of this section is to establish a fundamental understanding of these concepts which are indispensable for the comprehension of the subsequent analyses and discussions in this work.

We are concerned with the computational handling of textual opinion, sentiment, and subjectivity.
Our focus is on using computational methods to analyze and process subjective and objective information, such as opinions, attitudes, and emotions, in textual data.

\emph{Opinion mining} and \emph{sentiment analysis} (SA) refer to the identical topic of research in natural language processing (NLP) \citep{Pang2008, Liu_2012}. The initial step in sentiment analysis involves categorizing textual content as either subjective or objective. Objective text is grounded in verifiable facts and is independent of personal opinions or emotions. Conversely, subjective text expresses a personal opinion or feeling that is open to debate and not reliant on objective facts.
Thus, the primary emphasis lies on the task of \emph{subjectivity classification}. The second phase entails examining the subjective language and determining its sentiment polarity, such as whether it is positive, negative, or neutral. Sentiment analysis (SA) can be formally described as "Given a text $t$ from a text set $T$, computationally assigning polarity labels $p$ from a set of polarities $P$ in such a way that $p$ would reflect the actual polarity that is found in $t$." \citep{S.Alotaibi_2016}.

The level of granularity $t$ in textual analysis refers to the specific unit of study within a text, such as a term, phrase, sentence, or full document. The $t$-level sentiment analysis task is created by assigning a polarity label to this level. For instance,  document-level SA entails giving a single polarity label to encapsulate the sentiment expressed throughout an entire document. In recent times, research has expanded to encompass aspect level, which concentrates on assessing sentiment associated with particular elements or features mentioned in the text, hence giving rise to \emph{aspect-level} sentiment analysis.
This strategy is especially valuable in situations where a single text addresses many topics or conveys diverse opinions about different elements of a single topic.

The sentiment classes to be considered are determined by the set of polarities $P$. For instance, Binary SA (BSA) is the term used to describe sentiment analysis in the scenario where $P =\{positive, negative\}$. The sentiment classes for Ternary SA (TSA) applications now include a third value for neutral text, and the set of polarities is denoted as $P=\{positive, negative, neutral\}$.  An additional form of SA application pertains to emotions. For instance, the collection of polarities is denoted as $P=\{anger, disgust, fear, sadness, surprise, joy, love\}$. The task is transformed into a multiple-class SA in this instance.

In SA, considering the context of the examined text is essential because a text might have varying sentiments depending on the context. The context of a text encompasses various elements, including the author's background, the temporal and spatial setting in which the work was produced, and any cultural or social aspects that may impact its interpretation.

The Arabic language is abundantly diverse, with Modern Standard Arabic (MSA) in formal settings and a variety of dialects that are spoken in the Middle East and North Africa. This linguistic variety presents substantial obstacles to SA, as the same words or phrases can convey distinct sentiments in various dialects.  It is essential to integrate contextual and cultural understanding into SA models to effectively address these challenges.  This contextual sensitivity is essential for the creation of sentiment analysis tools that are more accurate and sophisticated for the Arabic language.

Finally, SA has a diverse array of applications in a variety of domains, including but not limited to social media monitoring, political analysis, financial analysis, hotel and restaurant review analysis, and movie reviews. SA offers these sectors the opportunity to gain a more comprehensive understanding of consumer behavior, public opinion, and the general sentiment trends that are pertinent to their respective fields.

\section{Related Work}
\label{sec:related}

We provide an overview of previous studies conducted on subjectivity and sentiment classification, with a specific focus on the English and Arabic languages. The works are classified into three categories: traditional, machine learning, and deep learning approaches.

\subsection{Subjectivity and Sentiment Classification for the English Language}

A variety of techniques and approaches are employed to handle the issue of text subjectivity classification. The English language is their primary focus. 

\subsubsection{Traditional Approaches}
Traditional methods employ manually designed features, including lexicon-based techniques, part-of-speech (POS) tagging, and dependency parsing. Lexicon-based approaches are based on precompiled collections of words and phrases, termed sentiment lexicons or subjectivity lexicons, such as SentiWordNet~\citep{Liu_2012}. Each term in the lexicon is assigned a score or label that denotes its level of subjectivity. Lexicon-based methods have produced positive results in some applications of sentiment classification. The authors in \cite{Kouloumpis_Wilson_Moore_2021} use a variety of features, including lexicon, POS, and microblogging features. The experiments carried out on Twitter sentiment analysis demonstrate that POS features may not be beneficial for sentiment analysis in the microblogging domain, and the lexicon features are moderately useful when used in combination with microblogging features. The paper \citep{10.5555/1699648.1699700} focuses on the task of mining subjective information from product reviews. The concept of phrase dependency parsing is presented using the observation that many product features are expressed as phrases. This concept extends the classic dependency parsing to the level of phrases. Empirical assessments demonstrate that the mining task can get advantages from phrase dependency parsing.

\subsubsection{Machine Learning Approaches}
English subjectivity classification has seen the application of machine learning approaches in recent years. The authors of \cite{pang-etal-2002-thumbs} consider the problem of classifying documents based on their general sentiment rather than their topic. The researchers employ three machine learning algorithms, namely SVM, NB, and maximum entropy classification, to analyze movie reviews. It has been discovered that the employed machine learning algorithms surpass human-produced baselines. Nevertheless, the applied algorithms do not achieve the same level of performance in sentiment classification as they do in traditional topic-based categorization. They conclude that the task of sentiment classification is considerably more challenging. The authors address the issue of classifying the subjectivity of sentences in~\cite{10.5555/1788714.1788746}. Their methodology involves utilizing a semi-supervised learning technique and self-training to accurately categorize sentences as either subjective or objective. The authors employ decision tree models, namely C4.5, C4.4, and naive Bayes tree (NBTree), as the underlying classifiers to investigate the performance of self-training.  The findings indicate that the self-training strategy can attain a level of performance that is similar to that of the supervised learning models.

\subsubsection{Deep Learning Approaches}
Deep learning methods have since also been extensively applied for subjectivity classification~\citep{Habimana2019SentimentAU}. \cite{johnson-zhang-2017-deep} aim to tackle the task of text categorization by developing a method that can efficiently capture and express long-range associations in text. The authors introduce a word-level convolutional neural network (CNN) structure, termed deep pyramid CNN (DPCNN). The experiments conducted on eight datasets compiled by~\cite{NIPS2015_250cf8b5} demonstrate that the proposed model surpasses the top-performing previous models in six datasets for sentiment classification and topic categorization. The research carried out in~\cite{conneau-etal-2017-deep} specifically examines sentence classification tasks. The authors present a very deep convolutional neural network (VDCNN) architecture to analyze text at the character level. The experiments are performed on eight datasets collected by~\cite{NIPS2015_250cf8b5}. These datasets encompass several classification tasks such as sentiment analysis, topic classification, and news categorization. The results demonstrate that VDCNN outperforms previously developed models on all datasets.

In contrast to CNN models, Recurrent Neural Networks (RNNs) are inherently designed to process sequential data. This makes RNNs highly suitable for tasks involving sequential information, such as sentiment analysis. For instance, \cite{Chen2016NeuralSC} want to tackle the task of sentiment classification at the document level. The objective is to forecast the overall sentiment expressed by users in a document regarding a product. The authors propose a hierarchical LSTM model to incorporate user and product information as attention mechanisms in sentiment categorization. To validate their model, the researchers perform experiments on various real-world datasets containing user and product information. These datasets include IMDB, Yelp 2013, and Yelp 2014, which were constructed by \cite{tang-etal-2015-learning}. The results indicate that the proposed model surpasses existing cutting-edge models. The study described in \cite{Wang_Pan_Dahlmeier_Xiao_2017} deals with the task of extracting both aspect and opinion terms simultaneously. This work involves the explicit extraction of aspect terms, which are words or phrases that describe aspects of an entity, as well as opinion terms that reflect emotions, from user-generated texts. The researchers provide a coupled multi-layer attention network. The model obtains state-of-the-art performances as evidenced by experimental results on three benchmark datasets from the SemEval Challenge 2014 and 2015. The study conducted in~\cite{giannakopoulos-etal-2017-unsupervised} centers around the task of aspect term extraction (ATE), which involves identifying opinionated aspect terms in texts. ATE is a component of the SemEval Aspect Based Sentiment Analysis (ABSA). The scarcity of datasets for ATE requires the use of unsupervised ATE. The authors employ a two-layer Bidirectional Long-Short Term Memory (B-LSTM) in this particular circumstance. The model is assessed using the human datasets from SemEval 2014 ABSA. The proposed unsupervised technique outperforms the supervised ABSA baseline from SemEval.  In~\cite{ghosal-etal-2018-contextual}, the focus is on multi-modal sentiment analysis. The researchers propose a multi-modal RNN framework that utilizes contextual information to predict sentiment at the utterance level. The evaluation of the proposed approach uses two benchmark datasets, specifically the Multi-modal Opinion-level Sentiment Intensity (MOSI) and Multi-modal Opinion Sentiment and Emotion Intensity (MOSEI). The findings show that the proposed model performs better than various state-of-the-art models. In summary, all proposed RNN models perform well in sentiment analysis and subjectivity classification. The computational expense of training these models might be a major limitation, especially when dealing with large datasets or complex architectures. Fortunately, the issue has been significantly mitigated by advancements in GPU technology.

\subsection{Subjectivity and Sentiment Classification for the Arabic Language}

Similarly, when examining the subjectivity classification in the English language, we may classify research on subjectivity classification in Arabic into three main approaches: traditional, machine learning, and deep learning.

\subsubsection{Traditional Approaches} In accordance with traditional methods, the study in ~\cite{Awwad2016PerformanceCO} focuses on the use of a lexicon-based technique for sentiment analysis (SA) in Arabic, namely at the document-level and sentence-level. The objective of this work is to conduct comparative research of various lexicons for Arabic sentiment analysis, with a focus on performance.  First, the authors implement an unsupervised approach for SA to determine the document polarity in Arabic. They involve the comparison of four lexicons: a translation of the Harvard IV-4 Dictionary (HarvardA), a translation of the MPQA (Multi-Perspective Question Answering) subjectivity lexicon developed by Pittsburgh University (HRMA), and two different implementations of MPQA. 
The lexicons are assessed using three datasets from various domains: PatientJo (a health domain dataset gathered by the authors of \cite{Awwad2016PerformanceCO} from three Jordanian hospitals), TA (Twitter Dataset for Arabic Sentiment Analysis), and LABR (Large scale Arabic Book Reviews). Empirical studies have demonstrated that individuals are more inclined to communicate their negative encounters within the healthcare industry, while they are more likely to express their positive experiences in book reviews. Moreover, the results indicate that both the lexicon-based strategy for document-level methods and sentence-level methods yield comparable performance.

\cite{abdul-mageed-etal-2011-subjectivity} aim to fill the gap of the scarcity of systems dealing with subjectivity and sentiment analysis (SSA) for morphologically-rich languages (MRL). They first create a corpus of modern standard Arabic (MSA) that has been carefully annotated, along with a new polarity lexicon. The annotation is performed at the sentence level by two college-educated native Arabic speakers who have received a college education. Subsequently, they examine the influence of various levels of preprocessing settings on the SSA task. The researchers conduct experiments using the Penn Arabic Treebank (PATB) \citep{Maamouri2004ThePA} and integrate a combination of language-independent and Arabic-specific morphology-based features. The empirical findings show that incorporating language with specific features for MRL leads to enhanced performance. Furthermore, they demonstrate that utilizing a polarity lexicon has the most significant influence on performance. 

In ~\cite{abdul2014samar}, Abdul-Mageed et al. developed SAMAR, a subjectivity and sentiment analysis (SSA) tool designed specifically for Arabic social media. The researchers aim to address four key research questions: the most effective way to represent lexical information; the relevance of standard features used in English for Arabic analysis; strategies for handling Arabic dialects; and the potential impact of genre-specific features on performance. To accomplish this, they created annotated data consisting of multiple datasets: DARDASHA, TAGREED, TAHRIR, and MONTADA. The findings indicate that incorporating lemma or lexeme information, as well as utilizing both reduced tag set (RTS) and extended reduced tag set (ERTS), is beneficial. Nevertheless, the findings indicate that distinct solutions tailored to each genre and purpose are necessary, while lemmatization and the ERTS POS tagset are prevalent in most configurations.

\subsubsection{Machine Learning Approaches} While there have been several machine learning classifiers employed for Arabic Sentiment Analysis in the literature, only three consistently exhibited superior performance: SVM, KNN, and NB~\citep{OUESLATI2020408review}.

The research conducted by \cite{doi:10.1177/0165551514534143} focuses on the effects of preprocessing strategies on Arabic sentiment analysis. Firstly, the study explores various options for text representation.  Furthermore, an examination was conducted on the performance of three classifiers, namely SVM, Naïve Bayes, and K-nearest neighbor classifiers, concerning sentiment analysis. The experiments employ two datasets.  The initial dataset was created by manually gathering reviewers' opinions from the Aljazeera website regarding various published political pieces. The second corpus is an Arabic opinion corpus that has been made freely accessible for research purposes and was developed by \cite{10.1002/asi.21598}. The results indicate that the implemented preprocessing procedures improve the effectiveness of all three classifiers.

\subsubsection{Deep Learning Approaches} Recently, deep learning methods have been proposed for Arabic subjectivity and sentiment classification. The article referenced as \citep{Alhumoud2022} presents a comprehensive examination of 24 research articles that employ Recurrent Neural Networks (RNNs) for Arabic sentiment analysis. Additionally, the study introduces novel datasets specifically designed for Arabic language sentiment analysis. Various researchers employ distinct models, including LSTM, Bi-LSTM, GRU, and hybrid models. Hybridization incorporates CNN architectures. The experiments utilized several datasets, including but not limited to LABR, ASTD, ArTwitter, Qatar Computing Research Institute (QCRI), SemEval-2017 Task 4, SemEval-2018 Task 1, SemEval-2016 Task 7, and ArSAS. The overall consensus from these studies is that employing Recurrent Neural Networks (RNNs) in sentiment analysis has demonstrated effectiveness, as these networks excel in textual analysis.

More recently, a new wave of studies has taken advantage of the ensemble methods. \cite{Alharbi2021} propose a Deep Learning model for Arabic Sentiment Analysis (DeepASA) that consists of two types of recurrent neural networks, namely GRU and LSTM. The voting-based ensemble technique, which utilizes majority voting, takes the output from both networks as input. This ensemble technique consists of three machine learning classifiers that are used to predict the class of each document. The tests are carried out using six datasets: LABR, Hotel Reviews (HTL), Restaurant Reviews (RES), Product Reviews (PROD), Twitter Data Set (ArTwitter), and ASTD. DeepASA demonstrated superior performance compared to current state-of-the-art results across all datasets, resulting in a considerable reduction in the classification error rate. In~\cite{Mohamed2022}, the authors want to improve the robustness and the performance of Arabic sentiment analysis leveraging transformer technology. They provide an ensemble method that merges two advanced transformer models: MARBERT (monolingual) and XLM-T (multilingual). The experiments involve a variety of datasets, namely ASTD, ArSarcasm-v2, and SemEval-2017. The results show that the proposed ensemble learning strategy surpasses state-of-the-art models on all the used datasets.

At the end of the present section, it is worth noting that language models have recently made great progress in improving the accuracy of English text classification. This enhancement is accomplished by pre-training these models on a large dataset and subsequently fine-tuning them for specific downstream tasks. This two-step procedure utilizes the extensive knowledge gained from diverse linguistic contexts during pre-training and applies it to attain superior performance in targeted applications. \cite{Alduailej2022} present AraXLNet, a novel Arabic language model. The development of this model involved pre-training the state-of-the-art XLNet model using a substantial Arabic corpus, specifically the OpenSubtitles, HARD, LABR, and Books Reviews in Arabic Dataset (BRAD). Subsequently, the model undergoes fine-tuning using several annotated Twitter Arabic datasets specifically designed for sentiment analysis. These datasets are AraSenTi, SemEval-2017, Arabic Jordanian General Tweets (AJGT), and ASTD. According to the experimental findings, the proposed approach demonstrates encouraging advancements in Arabic text categorization problems.

\section{Method}
\label{sec:method}
This study aims to develop a robust tool for evaluating subjectivity and analyzing sentiment within Arabic textual data. To achieve this, we propose a transformer-based solution that consists of fine-tuning three state-of-the-art Arabic language models: XLM-RoBERTa~\citep{conneau-unsupervised-2020},  AraBERT~\citep{antoun2020arabert}, and ArabianGPT-01B~\citep{koubaa2024arabiangpt} on the downstream task of Arabic text subjectivity classification. Additionally, we explore an ensemble approach, utilizing a voting-based technique, to combine the strengths of these individual models and potentially enhance classification performance.  

To fine-tune our models, we compiled a comprehensive dataset by combining the following publicly available Arabic datasets: Arabic Sentiment Tweets Dataset (ASTD)~\citep{nabil-etal-2015-astd}, Large Arabic Books Reviews (LABR)~\citep{aly-atiya-2013-labr}, Hotel Arabic-Reviews Dataset (HARD)~\citep{ELNAGAR2018182AnAnnotated} and Single-labeled Arabic News Articles Dataset (SANAD)~\citep{einea2019sanad}. This combined dataset provided a diverse and representative corpus for training our models.

In the next subsections, we give more details on datasets, preprocessing steps, fine-tuning, and testing for the proposed solution. 

\subsection{Dataset preparation}

\par To build our Arabic subjectivity classification models, a large amount of labeled data is required for fine-tuning and testing. These data should be annotated carefully as either subjective or objective to be able to train and test our models. We compile a large corpus of texts with annotations, leveraging already existing datasets.
The process of creating this dataset, termed \textit{AraDhati+}, consists of the following steps:

\subsubsection{Data collection}

Apart from the ASTD \citep{nabil-etal-2015-astd} dataset, we are unaware of any other dataset in which texts are annotated as subjective or objective. All other encountered datasets (LABR and HARD) are dedicated to sentiment analysis and classification, in which subjective text is annotated as either positive, neutral, or negative. To overcome this, we consider any text having one of these labels as a subjective one. 
For the objective data, we opted to leverage the SANAD~\citep{einea2019sanad} dataset, which is a single labeled large collection of Arabic news articles categorized into one of the following classes: Culture, Finance, Medical, Politics, Religion, Sports and
Technology. Articles from the Medical, Sports, and Technology sections are typically regarded as providing objective viewpoints on facts relating to medical science, sports activities and contests, and technology developments, respectively.

In the following paragraphs, we briefly describe each dataset used for building our training and testing dataset.
\begin{description}
    \item[\textbf{ASTD}] (Arabic Sentiment Tweets Dataset\footnote{ASTD is free and publicly accessible at:\url{https://github.com/mahmoudnabil/ASTD}})\,\citep{nabil-etal-2015-astd} is a dataset for Arabic social sentiment analysis sourced from Twitter. It comprises approximately $10,000$ Tweets classified into objective, subjective positive, subjective negative, and subjective mixed.

    \item[\textbf{LABR}] (Large-Scale Arabic Book Reviews\footnote{LABR is free and publicly accessible at:\url{https://github.com/mohamedadaly/LABR}})\,\citep{aly-atiya-2013-labr} is an Arabic large sentiment analysis dataset, consisting of over $63,000$ book reviews, each rated on a scale of 1 to 5 where ratings of 4 or 5 were considered positive, ratings of 1 or 2 were considered negative, and a rating of 3 is considered neutral.
    
    \item[\textbf{HARD}] (Hotel Arabic-Reviews Data set\footnote{HARD is free and publicly accessible at:\url{https://github.com/elnagara/HARD-Arabic-Dataset}})\,\citep{ELNAGAR2018182AnAnnotated} is an Arabic data set, comprising $490,587$ hotel reviews collected from the Booking.com website, where the reviews are expressed in Modern Standard Arabic as well as dialectal Arabic. We considered the balanced version of HARD which consists of $94,052$ reviews, each review is rated on a scale of 1 to 5 stars divided into positive with ratings of  4 and  5 and negative with ratings of  1 and 2. However, the neutral reviews with a rating of 3 have been removed from this version of the data set.
    
    \item[\textbf{SANAD}] (Single-labeled Arabic News Articles Dataset\footnote{SANAD is free and publicly accessible at:\url{https://data.mendeley.com/datasets/57zpx667y9/2}})\,\citep{einea2019sanad} is a large Arabic data set of textual data consisting of  194,797 articles combined from three datasets that were extracted from three news sources, which are AlKhaleej, Akhbarona, and AlArabiya. Articles fall into one of seven categories: Medical, Finance, Culture, Politics, Religion, Sports and Technology.

\end{description}

\subsubsection{Data Balancing and Augmentation}

The ASTD is a Twitter-based Arabic social sentiment analysis collection. It contains approximately 10K Tweets categorized as objective, subjective positive, subjective negative, or subjective mixed. The three subjective classes are merged to have only two classes: objective Tweets vs subjective Tweets. Table~\ref{tab:ASTDStatistics} presents the statistics regarding ASTD. 

\begin{figure}[htbp]
    \centering
    \includegraphics[width=0.8\textwidth ]{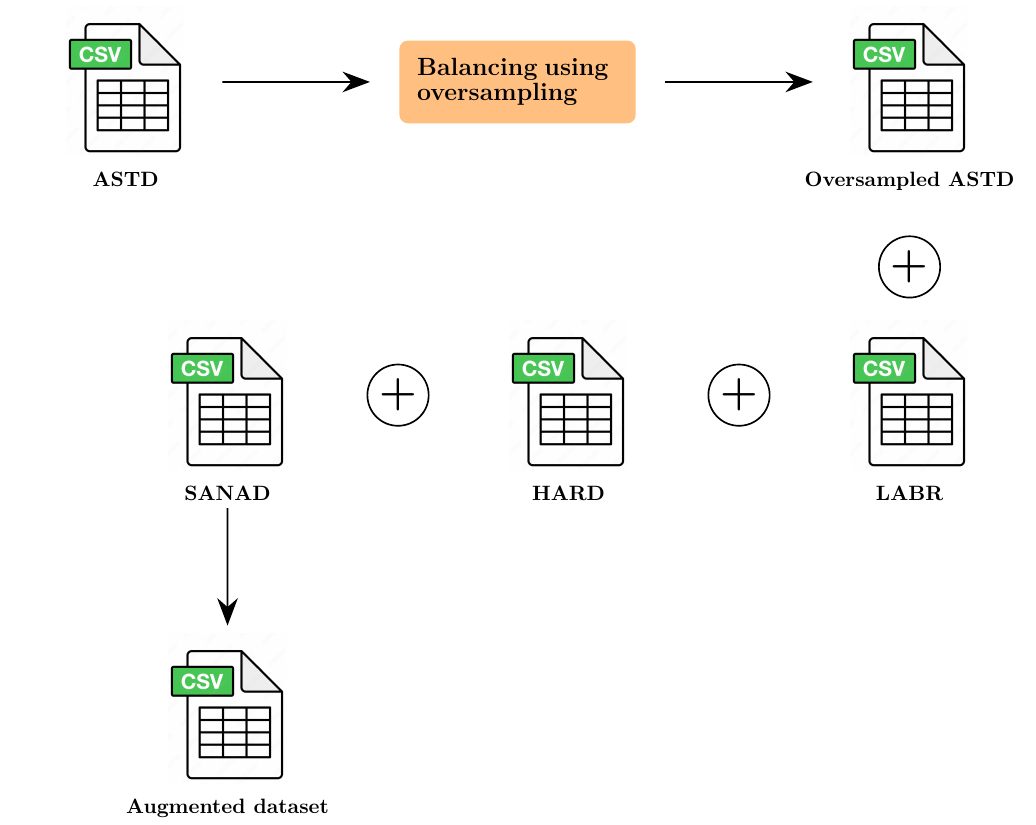}
    \caption{Data Balancing and Augmentation process}
    \label{fig:augment_data}
\end{figure}

Unfortunately, this dataset is unbalanced. Thus, we first explore an oversampling technique to have two classes of equal size. Oversampling consists of re-sampling (duplicating instances) from the underrepresented class. 
Then, to further enhance the dataset, we explore an augmentation technique, which consists of adding new instances from external resources. We incorporate the above-mentioned datasets: LABR, HARD, and SANAD. This decision was driven by the motivation to create a larger and more diverse dataset, and to expand the range of subjective and objective texts available for training models.
From the SANAD dataset, we extracted $32,500$ news articles which are considered objective texts. This addition was essential to introduce a substantial number of unbiased samples, allowing the models to learn the distinguishing features of objective language.
To maintain the balance of the augmented dataset, we selected $32,500$ subjective reviews equally picket from the LABR and HARD datasets. These datasets were carefully chosen for their relevance and suitability to the subjective language.  Figure~\ref{fig:augment_data} provides a visual representation of the data balancing and augmentation process for easy reference.

By incorporating these additional datasets, we aimed to achieve two main objectives. First, we sought to increase the overall size of the over-sampled version of ASTD, providing a more comprehensive and diverse training set for models. Second, we aimed to keep a balance between subjective and objective instances within the augmented dataset, ensuring that models receive adequate exposure to both types of texts.

\subsubsection{Dataset cleaning and Normalization}

To improve the dataset quality, we cleaned up the unwanted content by removing non-useful text like URLs, non-Arabic characters, punctuation marks, special characters, single letters, etc. Normalization is also applied to words. Removing stop words is not convenient in this context because it can affect the assessment of subjectivity and sentiment analysis.

\begin{table}[tbp]
	\centering
 	\caption{ASTD Statistics \citep{nabil-etal-2015-astd}}
		\begin{tabular}{|c|c|}
  \hline
			Total number of Tweets & 10,006 \\\hline
                Subjective Tweets & 3,315 \\\hline
                Objective Tweets & 6,691 \\\hline
                Max tokens per Tweet & 45 \\\hline
                Avg. tokens per Tweet & 16 \\\hline
                Number of tokens & 160,206 \\\hline
                Vocabulary size & 38,743 \\\hline
   
		\end{tabular}
	\label{tab:ASTDStatistics}
\end{table}

\subsubsection{Dataset Formating}
The augmented data file is formatted in a commonly used and easily accessible CSV (Comma-Separated Values) format. It is organized in rows and columns, where each row corresponds to a unique text instance, and each column represents a specific attribute. Columns are as follows:

\begin{itemize}
    \item \textbf{Text:} holds the actual text data.
    
    \item \textbf{Class:} indicates the subjectivity class for each instance. The subjectivity classes are categorized as \textit{POS} (positive), \textit{NEG} (negative), \textit{NEUTRAL}, and \textit{OBJ} (objective).

    \item \textbf{Domain:} specifies the domain of the text, such as "Tweets", "Book reviews", "Hotel reviews", or "Sports".

    \item \textbf{Label:} gives a numerical representation of subjectivity, with 1 denoting subjective text and 0 denoting objective text.
    
    \item \textbf{Dataset:} indicates the original dataset from which text was picked.
\end{itemize}

Table~\ref{tab:Aug_data} provides instance examples of the augmented dataset. It showcases the different columns and their corresponding values, offering a clear representation of the augmented dataset.


\begin{table*}[htbp]
\fontsize{8}{9}\selectfont
\centering
    \caption{Examples from the balanced and augmented dataset}
\vspace*{2mm}
\begin{tabular}{|>{\centering\arraybackslash}p{7cm}|c|c|c|c|}
\hline
Text & Class & Domain & Label & Dataset \\
\hline
\RL{أهنئ الدكتور أحمد جمال الدين، القيادي بحزب مصر، بمناسبة صدور أولى روايته} & POS & Tweets & 1 & ASTD \\
\hline
\RL{فعلا نصائح مميزة ومفيدة جدا احسن حاجة في الأحلام أنها بتكتب عن تجربة} & POS & Books reviews & 1 & LABR \\
\hline
\RL{فندق فاشل. انا حجزت ووصلت فالموعد ولم اجد غرف. ما يصلح فاشل النظام} & NEG & Hotel reviews & 1 & HARD \\
\hline
\begin{RLtext}أعلنت شركة جوتن، إحدى أبرز الشركات العالمية في مجال إنتاج وتوريد الدهان والطلاء وبودرة الطلاء، عن إطلاقها النسخة العربية من موقعها الإلكتروني\end{RLtext} & OBJ & Technology & 0 & SANAD \\
\hline
\end{tabular}
\label{tab:Aug_data}
\end{table*}

\subsubsection{Dataset Splitting}
Typically, the collected dataset is split into training and testing sets. The training set, comprising $80\,\%$ of the entire collection, is used to train the models, while the remaining $20\,\%$ is reserved for testing to evaluate their performance. We ensure class balancing and a random split to maintain a representative data distribution. Table~\ref{tab:train-test2} provides statistics on the training and testing parts of our AraDhati+ dataset. The dataset is publicly available \footnote{\url{https://github.com/Attia14/AraDhati}}.

 \begin{table}[htbp]
    \centering
    \caption{Statistics of the training and the testing sets}
		\begin{tabular} {|c|c|c|}
			\hline 
			& Train data 	& 	Test data \\  \hline 
			 ASTD & $10,332$ &  $2,584$  \\ 
			\hline
             LABR & $13,000$ & $3,250$ \\  
             \hline
             HARD & $13,000$ & $3,250$ \\ 
             \hline
             SANAD & $26,000$ & $6,500$ \\ 
             \hline
             Total & $62,332$ & $15,584$ \\  
             \hline
		\end{tabular}
		\label{tab:train-test2}
	\end{table}

\subsection{Fine-tuning pre-trained models}

In our approach, we fine-tune pre-trained language models for the downstream task of Arabic text subjectivity classification and evaluate their performance. We use the Hugging Face Transformers library to fine-tune the XLM-RoBERTa, AraBERT, and ArabianGPT on the Arabic subjectivity classification task using the training set of our augmented dataset.

In the following paragraphs, we provide a succinct description of each model along with the details of each scenario:

\subsubsection{XLM-RoBERTa}
The first used model is XLM-RoBERTa, which is a multilingual model based on  RoBERTa (a transformer model pre-trained on a large corpus in a self-supervised fashion), and it was pre-trained with the Masked Language Modeling (MLM) objective on 2.5\, TB of filtered CommonCrawl data containing 100 languages including Arabic. XLM-RoBERTa uses the same MLM objective as the XLM model with only one change: removing the language embeddings, allowing the model to better deal with code-switching. This way, the model can learn useful representations of languages to produce helpful features for specific tasks~\citep{conneau2019unsupervised}.
One of the advantages of XLM-RoBERTa's multilingual pre-training is that it can be fine-tuned for specific downstream tasks, such as text classification.

\begin{figure}[htbp]
    \centering
    \includegraphics[width=0.9\textwidth ]{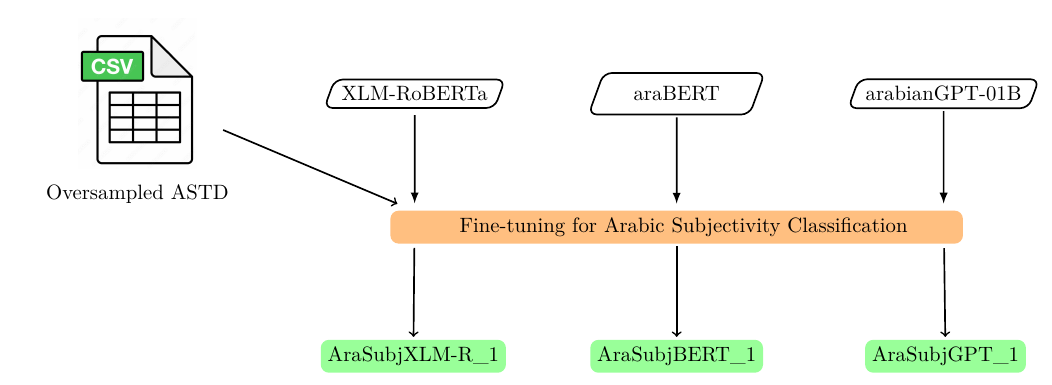}
    \caption{Fine-tuning XLM-RoBERTa, araBERT, and arabianGPT-01B using the oversampled ASTD}
    \label{fig:finetuning-using-dhati}
\end{figure}

Initially, we fine-tuned XLM-RoBERTa on the over-sampled version of the ASTD dataset, which was balanced to address the class imbalance problem. Figure~\ref{fig:finetuning-using-dhati} visually represents the fine-tuning process on the over-sampled ASTD. 
The resulting model is subsequently referred to as AraSubjXLM-R\_1. 

\begin{figure}[htbp]
    \centering
    \includegraphics[width=0.9\textwidth ]{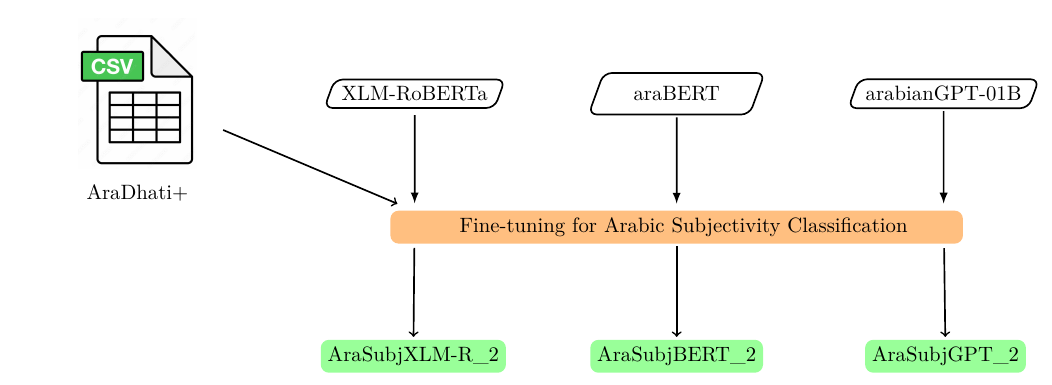}
    \caption{Fine-tuning XLM-RoBERTa, araBERT, and arabianGPT-01B using AraDhati+ (the augmented ASTD)}
    \label{fig:finetunig-using-dhatiplus}
\end{figure}

Then, to determine the impact of the additional data on the model's accuracy, we fine-tuned XLM-RoBERTa on the augmented dataset. Figure~\ref{fig:finetunig-using-dhatiplus} provides an overview of this process. 
The resulting model is referred to as AraSubjXLM-R\_2.

\subsubsection{AraBERT}
The second model we fine-tuned is AraBERT, which is an Arabic language representation model based on the BERT~\citep{devlin2018bert} LLM (a stacked Bidirectional Encoder Representations from Transformer), that has been pre-trained particularly for Arabic to accomplish the successful that BERT achieved for English. 
\newline Basically, BERT was trained on 3.3 billion words from English Wikipedia and Book Corpus. However, due to the lack of Arabic resources, AraBERT was trained on 70 million sentences extracted from two large corpora, the 1.5 billion words Arabic Corpus and the Open Source International Arabic News Corpus (OSIAN).
AraBERT architecture is based on the original BERT architecture but specifically designed and trained for the Arabic language. It comprises 12 encoder blocks, 768 hidden dimensions, 12 attention heads, 512 maximum sequence length, and 136M parameters~\citep{antoun2020arabert}.

We fine-tuned AraBERT on the oversampled and augmented versions of the ASTD. The resulting models are referred to as AraSubjBERT\_1 and AraSubjBERT\_2, respectively (Figures~\ref{fig:finetuning-using-dhati} and \ref{fig:finetunig-using-dhatiplus}).

\subsubsection{ArabianGPT}
The third model we fine-tuned is ArabianGPT-01B~\citep{koubaa2024arabiangpt}, which is developed as part of the ArabianLLM projects. This model builds upon the GPT-2 architecture, specifically optimized for the intricacies of the Arabic language. The collaborative efforts behind ArabianGPT-0.1B stem from Prince Sultan University's Robotics and Internet of Things Lab. Their primary focus lies in advancing the capabilities of Arabic language modeling and generation. Notably, ArabianGPT-0.1B represents a significant breakthrough in LLM research, tackling the unique linguistic features and subtleties inherent to Arabic. 

ArabianGPT-01B was trained on a 15.5GB dataset containing 237.8 million words from scraped Arabic newspaper articles. 
It uses the same architecture as GPT-2. The authors keep the GPT model's core ideas while adjusting (particularly the tokenizer) to suit Arabic text processing better. ArabianGPT-01B is specifically designed to understand and generate genuine Arabic text. It starts with an embedding layer and has 12 identical layers, each with three sub-layers: masked multi-head self-attention, a feed-forward network, and layer normalization. This stacking architecture enables the model to understand complicated links among words in a sentence. The model has 134M parameters~\citep{koubaa2024arabiangpt}.

As for the previous two models, we fine-tuned ArabianGPT on the oversampled and augmented versions of the ASTD. The resulting models are referred to as AraSubjGPT\_1 and AraSubjGPT\_2, respectively (Figures~\ref{fig:finetuning-using-dhati} and \ref{fig:finetunig-using-dhatiplus}).

\section{Experiments and Result Discussion}
\label{sec:experiments}

First, we introduce the environment in which our proposed solution was implemented. Next, we discuss the results of fine-tuning models on the oversampled and augmented versions of the ASTD.

\subsection{Configurations and Results}
 
Our solution was implemented using several libraries, including \textit{transformers}, \textit{imbalanced-learn}, \textit{pandas}, \textit{numpy}, and \textit{torch}.

For each model, we used the corresponding tokenizer (\textit{XLMRobertaTokenizerFast},  \textit{BertTokenizerFast}, and \textit{AraNizer}) to encode Tweets. Truncation and padding are used with a maximum length set to $256$. 
Fine-tuning was performed on a GPU platform using Mini-Batch Gradient Descent (MGD) with a mini-batch size of $16$ and the AdamW optimizer \citep{loshchilov2018decoupled} with various learning rates and 
epoch values. Table~\ref{tab:configs} summarizes configurations used in fine-tuning.  
Obtained models are evaluated using a hold-out method, a stratified testing subset from each dataset is used, with accuracy, precision, recall, and F1-score metrics. 

To assess the generalization capabilities of AraSubjXLM-R\_1, AraSubjBERT\_1, and AraSubjGPT\_1, which were fine-tuned on an oversampled version of the ASTD dataset, we conducted evaluations on the test set parts of the LABR, HARD, and SANAD datasets.
 
\begin{table}[tbp]
	\centering
 \caption{Configuration used in Fine-tuning XLM-RoBERTa, AraBERT, and arabianGPT2}
		\begin{tabular}{|c|c|}
  \hline
			Optimization Method & Mini Batch Gradient Descent \\\hline
            Optimizer & AdamW \\\hline
            Mini Batch size & 16 \\\hline
                Learning rate values & 5e-6 , 15e-6, 20e-6, 5e-5 \\\hline
                epochs & $\{1, 2, 3, 5, 7\}$ \\\hline       
		\end{tabular}
		\label{tab:configs}
\end{table}

\subsection{Discussion}
\begin{figure*}[htbp]
   \centering 
    \includegraphics{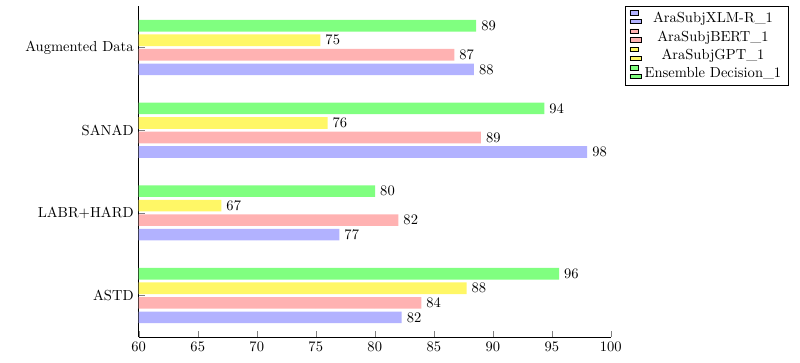}
    \caption{Performances of AraSubjXLM-R\_1,  AraSubjBERT-R\_1, AraSubjGPT\_1, and Ensemble Decision\_1 on all datasets}
\label{fig:performance-oversampled}
\end{figure*}

From results shown in Figure~\ref{fig:performance-oversampled}, we can draw general conclusions as follows: 
\begin{itemize}
    \item When comparing individual models, the AraSubjGPT\_1 exhibited the highest overall accuracy (87.78\%) on the ASTD test set. However, AraSubjXLM-R\_1 demonstrated superior generalization capabilities, achieving a remarkable 98\% accuracy on the objective SANAD test set. Conversely, AraSubjBERT\_1 showed stronger performance on subjective data, attaining an 82\% accuracy on the LABR-HARD benchmark. 

    \item The ensemble model Decision\_1, which was constructed to leverage the strengths of the individual models, demonstrated superior performance, achieving accuracies of 95.62\% and 88.60\% on the ASTD and augmented test sets, respectively. Notably, it maintained competitive performance on the subjective and objective external data parts with accuracies of 80.03\% and 94.37\%.
\end{itemize}

\begin{figure*}[htbp]
   \centering
      \includegraphics{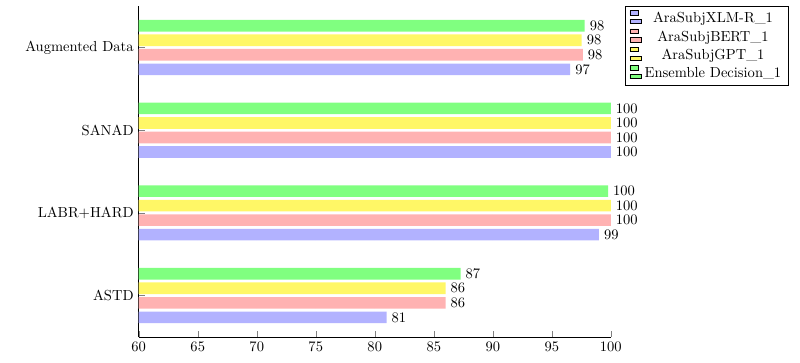}
   \caption{Performances of AraSubjXLM-R\_2,  AraSubjBERT-R\_2, AraSubjGPT\_2, and Ensemble Decision\_2 on all datasets}
\label{fig:performance-augmented}
\end{figure*}

From results shown in Figure~\ref{fig:performance-augmented}, we can draw general conclusions as follows: 

\begin{itemize}
    \item When comparing individual models, the AraSubjGPT\_2 and ArabSubjBERT\_2 exhibited the highest overall accuracy (86\%) on the ASTD test set. However, all models achieved remarkable accuracy on the objective and subjective test sets (more than~99\%). 
    
    \item Ensemble model Decision\_2 exhibited slightly improved performance compared to its components, attaining accuracies of 87.27\% and 97.79\% on the ASTD and augmented test sets, respectively.

    \item A comparative analysis of models trained on the oversampled ASTD versus the augmented dataset revealed enhanced performance on the objective and subjective test sets for the latter group. This improvement is attributed to the inclusion of additional data from SANAD and LABR-HARD during fine-tuning. Conversely, all models experienced a decline in performance, ranging from 1.28\% to 1.78\%, on the original ASTD test set. The observed decrease in performance on the original ASTD test set can be attributed to the phenomenon of domain shift. The models implicitly learn new distribution-specific patterns by fine-tuning them on the augmented dataset, which incorporates data from SANAD and LABR-HARD.  
    Consequently, a performance decline is observed when evaluated on the original ASTD, which follows a different data distribution. This underscores the challenges of adapting models to varying data distributions and domains. We conducted a detailed error analysis of the Ensemble Decision\_2 to understand these models' behavior better. 
\end{itemize}

\subsubsection*{Detailed error analysis}

\begin{figure*}[htbp]
   \centering
   
    \includegraphics[width=9cm,height=6cm]{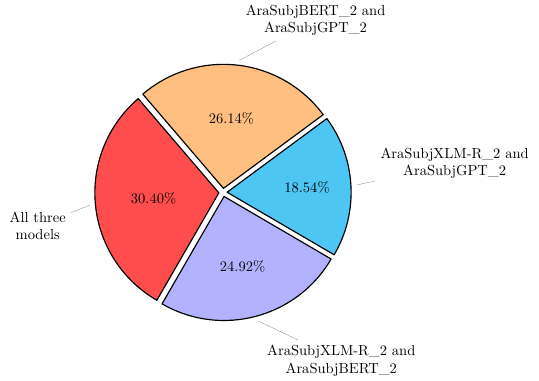}
    \caption{Distribution of the Ensemble Decision\_2 errors}
\label{fig:errors1}
\end{figure*}

Errors from the Ensemble Decision\_2 (Figure~\ref{fig:errors1}) can be categorized into two groups: those where only two components misclassified the text and those where all three components misclassified the text. We focused on the latter type of error (which represents 30.40\% of the misclassified instances) as understanding these instances provides valuable insights into the model's limitations. Indeed, when all components agree on an incorrect classification, it suggests a systematic flaw in the model's design or the underlying data. By analyzing these cases, we can identify potential areas for improvement and enhance the model's overall performance.

\begin{figure*}[htbp]
   \centering
    \includegraphics[width=8cm,height=5cm]{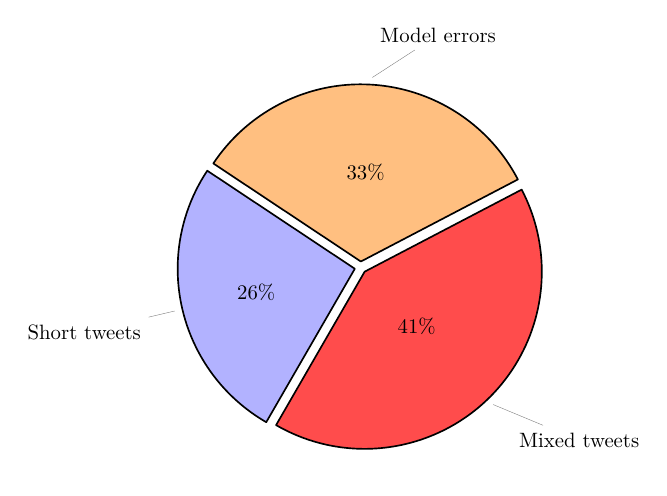}
    \caption{Categories of errors made by all three components of the Ensemble Decision\_2}
\label{fig:errors2}
\end{figure*}

A systematic analysis of the misclassified instances identified three distinct categories (as shown in Figure~\ref{fig:errors2}): (i) Mixed Tweets (41\,\% of the total errors), characterized by a combination of subjective and objective sentiment fragments within the same text; (ii) Model errors (33\,\% of the total errors), arising from the model's inherent limitations or biases; and (iii) Short Tweets (26\,\% of the total errors), which often lack sufficient context for accurate classification. Table~\ref{tab:misclassifed} provides representative examples from each category.

The first category, Mixed Tweets, consists of sentences that are neither entirely subjective nor entirely objective but rather a blend of both. This is a common phenomenon in natural language, where personal opinions are often expressed based on factual observations. As illustrated in Table~\ref{tab:misclassifed}, the first sentence exemplifies this. It begins with a factual assertion, "Competition is a natural thing," followed by a subjective opinion, "but its negative aspect is that it divides the voices of those close to each other in their ideas." While the initial statement is generally accepted as a fact, the subsequent opinion reflects a personal perspective on the potential consequences of competition, which may vary across individuals.

\begin{table*}[h]
\centering
\caption{Examples of misclassified instance by the Ensemble Decision\_2}
\vspace*{2mm}
\begin{tabular}{|>{\centering\arraybackslash}p{5cm}|p{5cm}|c|}
\hline
Misclassified instance & English translation & Category  \\
\hline
\begin{RLtext}المنافسة مسألة طبيعية، جانبها السلبي هو تفتيت الأصوات بين القريبين من بعض في أفكارهم\end{RLtext} & Competition is a natural thing, but its negative aspect is that it divides the voices of those close to each other in their ideas. & Mixed Tweets \\
\hline
\RL{الوفد - تأسيس السيسى لحزب سيصيب الحياة السياسية بالخلل} & \#Al-Wafd - Sisi's establishment of a party will disrupt political life & Model error \\
\hline
\RL{انا مبحبش الناس الأوفر انا مبحبش الناس عموما !! } & I don't like rude people, I don't like people in general!!  & Model error  \\
\hline

\RL{السعاده } & Happiness & Short Tweets  \\
\hline
\end{tabular}
\label{tab:misclassifed}
\end{table*}

The second category, Model Errors, encompasses sentences that were misclassified by the Ensemble Decision\_2. As illustrated in Table~\ref{tab:misclassifed}, the second and third sentences exemplify two subjective sentences that were erroneously classified as objective. The second sentence was mistakenly categorized as objective because it was perceived as verifiable news. However, it's often challenging to pinpoint the exact reasons behind the model's misclassifications, as demonstrated by the third sentence ("I don't like rude people, I don't like people in general!!"), which remains difficult to explain.

The third category, Short Tweets, consists of sentences that are too concise to provide adequate context for accurate classification. As illustrated in Table~\ref{tab:misclassifed}, the fourth sentence, "Happiness," is a prime example of this. Due to its brevity, it lacks the necessary linguistic cues for the model to effectively determine its sentiment.

In conclusion, our examination of the misclassified instances revealed three distinct categories: Mixed Tweets, Model Errors, and Short Tweets. Each category presents unique challenges for subjectivity analysis, highlighting the need for continual improvement in model architectures and training data.

\section{Conclusion}
\label{sec:conc}

Sentiment and subjectivity analysis tools face numerous challenges to reach accurate interpretation. A critical factor influencing the effectiveness of these tools is the contextual understanding of the text. Contextual elements, such as background information and surrounding text, can significantly impact the meaning and sentiment conveyed. Transformers provide an alternate method to conventional machine learning methods. 

In this work, we have focused on assessing subjectivity in Arabic textual data. We presented a solution to this problem by augmenting the Arabic Tweets dataset (ASTD) with external data from subjective and objective data sources, resulting in the creation of our AraDathi+ dataset. Subsequently, we fine-tuned three LLMs on this enhanced dataset. 

Our findings underscore the effectiveness of ensemble models in leveraging the complementary strengths of individual models. By combining diverse models, we achieved a peak performance of 97.79\%. However, the importance of data diversity cannot be overstated, as models trained on a broader spectrum of data demonstrated enhanced generalizability.

Future research should address the limitations identified in our experiments to further improve the accuracy and robustness of subjectivity and sentiment analysis models. This includes developing strategies to mitigate model drift and maintain performance over time. By prioritizing these areas, we can contribute to the advancement of sentiment analysis techniques for short and complex text.

\backmatter

\bmhead{Acknowledgements}
This work is supported by the General Direction of Scientific Research and Technological Development (DGRSDT) - Algeria, and performed under the PRFU Projects: C00L07N030120220002 and C00L07UN470120230001.

\bibliography{sn-bibliography}

\end{document}